\documentclass[letterpaper, 10pt, conference]{ieeeconf}      

\IEEEoverridecommandlockouts                              

\usepackage{amsmath,amssymb,amsfonts}
\usepackage{graphicx}
\usepackage{textcomp}
\usepackage{graphics} 
\usepackage{epsfig} 
\usepackage{subfigure}
\usepackage{amsmath} 
\usepackage{amssymb}  
\usepackage{epstopdf}
\usepackage{lipsum}
\usepackage{lipsum,amsmath,multicol}
\usepackage{float}
\usepackage{algorithm}
\usepackage{algpseudocode}
\usepackage{wrapfig}
\usepackage{sidecap}

\usepackage{amsthm}

\usepackage{comment}
\usepackage{array}
\usepackage{glossaries}
\usepackage{breqn}
\makeglossaries
\newcolumntype{L}[1]{>{\raggedright\let\newline\\\arraybackslash\hspace{0pt}}m{#1}}
\newcolumntype{C}[1]{>{\centering\let\newline\\\arraybackslash\hspace{0pt}}m{#1}}
\newcolumntype{R}[1]{>{\raggedleft\let\newline\\\arraybackslash\hspace{0pt}}m{#1}}
\usepackage{sidecap}
\usepackage{url}
\usepackage[T1]{fontenc}
\usepackage{tabularx}
\usepackage{multicol, blindtext}
\usepackage{amsmath}
\usepackage{multirow}
\usepackage{xcolor}
\usepackage{graphicx}
\usepackage{caption}

\usepackage{mathtools, bigstrut}
\usepackage{tabularx}
\usepackage{multicol, blindtext}
\usepackage{amsmath}
\usepackage{multirow}

\allowdisplaybreaks

\newcommand\blfootnote[1]{%
  \begingroup
  \renewcommand\thefootnote{}\footnote{#1}%
  \addtocounter{footnote}{-1}%
  \endgroup
}

\begin{document} 

\let\ACMmaketitle=\maketitle

\title{GPU Accelerated Batch Multi-Convex Trajectory Optimization for a Rectangular Holonomic Mobile Robot\footnote{Just a note.}}
\author{Fatemeh Rastgar, Houman Masnavi, Karl Kruusamäe, Alvo Aabloo, Arun Kumar Singh }

\twocolumn[{%
\renewcommand\twocolumn[1][]{#1}%
\maketitle
 \begin{center}
    \centering
  \includegraphics[scale=0.5]{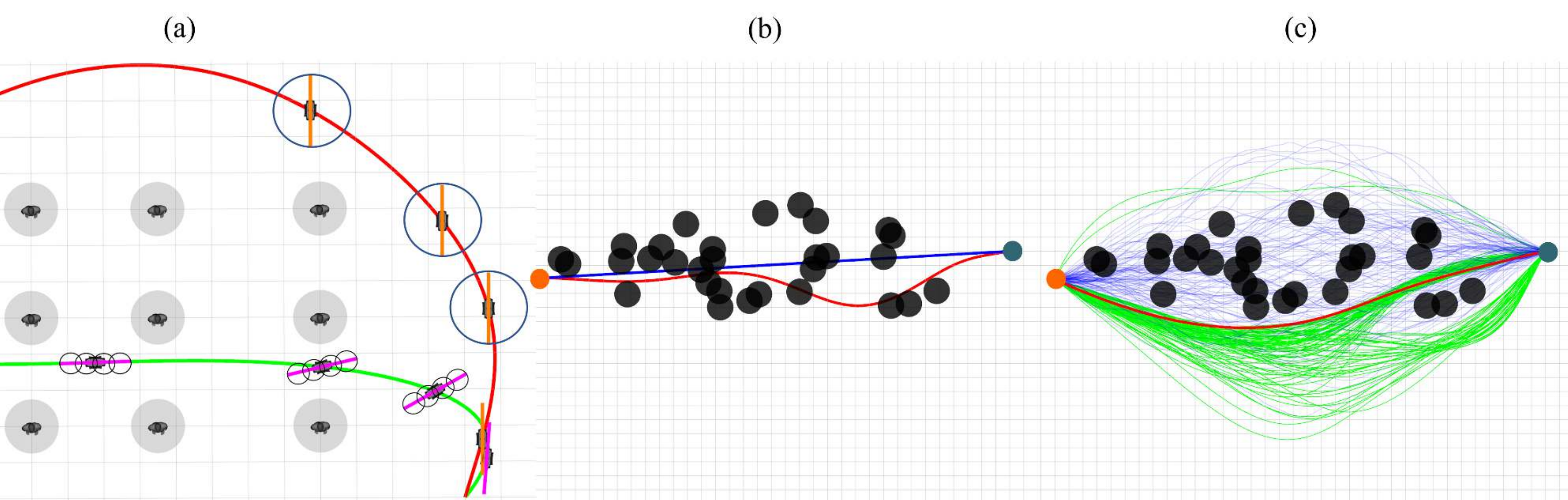}
\label{teaser}
\vspace{-0.2cm}
\captionof{figure}{(a): Robots with rectangular footprints are better modeled through a combination of overlapping circles. This allows for leveraging rotational motions to fit into tight spaces. In contrast, the circular footprints prove too conservative and may force the robot to take larger detours. \textbf{Our work makes trajectory optimization with multi-circle approximation more tractable.}(b): Naive initialization (e.g., straight blue line) may lead the trajectory optimizer to unsafe local minima. Our batch setting allows us to run tens of hundreds of different instances of the problem in real-time, obtained by different initialization of the problem. In (c), the blue trajectories represent initialization samples drawn from a Gaussian distribution \cite{stomp}. After few iterations, our batch optimizer returns a distribution of locally-optimal trajectories (green) residing in different homotopies (best cost trajectory: red)}
\vspace{-0.6cm}
\end{center}
}]

\blfootnote{All authors are with the Institute of Technology, University of Tartu. The work was supported in part by the European Social Fund through IT Academy program in Estonia, smart specialization project with BOLT and grants COVSG24 and PSG605 from Estonian Research Council.}





\begin{abstract}
We present a batch trajectory optimizer that can simultaneously solve hundreds of different instances of the problem in real-time. We consider holonomic robots but relax the assumption of circular base footprint. Our main algorithmic contributions lie in: (i) improving the computational tractability of the underlying non-convex problem and (ii) leveraging batch computation to mitigate initialization bottlenecks and improve solution quality. We achieve both goals by deriving a multi-convex reformulation of the kinematics and collision avoidance constraints. We exploit these structures through an Alternating Minimization approach and show that the resulting batch operation reduces to computing just matrix-vector products that can be trivially accelerated over GPUs. We improve the state-of-the-art in three respects. First, we improve quality of navigation (success-rate, tracking) as compared to baseline approach that relies on computing a single locally optimal trajectory at each control loop. Second, we show that when initialized with trajectory samples from a Gaussian distribution, our batch optimizer outperforms state-of-the-art cross-entropy method in solution quality. Finally, our batch optimizer is several orders of magnitude faster than the conceptually simpler alternative of running different optimization instances in parallel CPU threads. \textbf{Codes:} \url{https://tinyurl.com/a3b99m8}

\end{abstract}

\section{Introduction}
Mobile robots are becoming an attractive option for autonomous object transportation in malls, hospitals, airports, etc. Our current work stems from our effort to develop a fleet of autonomous delivery robots for hospital premises. We identify two key gaps in the existing literature on optimization-based motion planning and control in this context. First, the circular footprint assumption typically made for mobile robots becomes too conservative in object transportation set-up. For example, even if the base footprint of the robot is circular when an external object (e.g., racks in warehouses) is mounted on it, the footprint invariably takes an elongated rectangular shape. A straightforward workaround is to approximate the footprint through a combination of circles \cite{aude_billard_convex} (see Fig. 1(a) ). However, this dramatically increases the number of collision avoidance constraints making trajectory optimization more challenging as even the state-of-the-art interior-point optimizers scale cubically with the number of constraints.

The second challenge stems from the critical reliance of trajectory optimizer on initialization (see Fig. 1(b)). A poor initialization invariably compromises the collision avoidance behavior. We note that it is particularly challenging to develop good initialization in environments (e.g., hospitals) densely cluttered with dynamic obstacles whose trajectory can change rapidly. 

Our work addresses both the challenges discussed above, and we summarize our key contributions below.

\noindent\textbf{Algorithmic:} We present a linearization-free multi-convex approximation of the collision-avoidance, velocity, and acceleration constraints that results in an optimizer scaling linearly with the number of obstacles and circles used to approximate the rectangular footprint of the robot. Moreover, the optimizer can iterate over several instances of the problem in parallel (a.k.a the batch-setting). We show that we can leverage the batch operation to run the optimizer from different initialization in parallel, obtaining solutions in different homotopies and mitigating local minima traps. 


\noindent \textbf{State-of-the-art Performance:} We are not aware of any approach that can solve hundreds of different instances of non-convex trajectory optimization problems in parallel. Our work is a first such attempt in this direction. As we show later, the simpler alternative of running different instances on parallel CPU threads is not scalable to large batch sizes.  We perform extensive simulations in Model Predictive Control (MPC) setting and show that our batch optimizer is crucial for navigating highly cluttered and dynamic environments with tens of moving obstacles. More concretely, our batch MPC dramatically improves the success rate and solution quality compared to the baseline approach that computes only a single locally optimal trajectory at each control loop. We also benchmark our batch optimizer against the state-of-the-art Cross-Entropy Method (CEM) \cite{cem_1}, \cite{cem_2} that can also search over different local minima. Our approach achieves a higher success rate of collision avoidance and better trajectory-tracking than CEM. 

\section{Problem Formulation and Related Works}
\subsection{Symbols and Notations}
\noindent Small case regular and bold font letters represent scalars and vectors, respectively. Matrices have upper case bold fonts. $\mathbb{R}^{q}$ denotes the $q$-dimensional Euclidean space. Superscript $T$ shows the transpose of vectors and matrices, while $t$ represents the time stamp. The left superscript $k$ and subscript $l$ denote our trajectory optimizer's iteration and batch index. The main symbols are summarized in Table \ref{symbols} while the rest will be defined in their first place of use. 

\small
\begin{table}[!t]
\centering
\caption{Important Symbols  } \label{symbols}
\scriptsize
\begin{tabular}{|p{1.8cm}|p{0.45cm}|p{5cm}|p{3cm}|}\hline
\mbox{$q$, $n$, $m$, $n_{v}$} & Scalar & Planning steps, number of the obstacles, number of circles to model robot, number of optimization variable, respectively.
\\ \hline
\mbox{ $\boldsymbol{\lambda}_{l} $} & $\mathbb{R}^{4n_{v}}$ & Lagrangian multipliers
\\ \hline
\mbox{ $\boldsymbol{\lambda}_{\psi,l}$} & 
$\mathbb{R}^{n_{v}}$ & Heading angle Lagrangian multipliers
\\ \hline
\mbox{$\textbf{x}_{l}, \textbf{y}_{l}, \boldsymbol{\psi}_{l}$} & $\mathbb{R}^{q}$ & Position and heading angle\\ \hline
\mbox{$\boldsymbol{d_{v,l}}$, $\boldsymbol{d_{a,l}},\boldsymbol{d_{ij,l}}$}
\mbox{$\boldsymbol{\alpha_{v,l}}$, $\boldsymbol{\alpha_{a,l}}, \boldsymbol{\alpha_{ij,l}}$}
 &  $\mathbb{R}^{q}$ & Velocity and acceleration variables associated with Kinematics and collision avoidance constraints.\\ \hline
\mbox{$\boldsymbol{x_{j}}$, $\boldsymbol{y_{j}}$ }& $\mathbb{R}^{q}$  &  Position of the j-th obstacle
\\ \hline
\end{tabular}
\normalsize
\vspace{-0.7cm}
\end{table}
\normalsize

\subsection{Batch Trajectory Optimization for Rectangular Holonomic Robots }
\noindent We are interested in solving $l$ trajectory optimizations in parallel each of which has the following form.

\vspace{-0.4cm}
\small
\begin{subequations}
\begin{align}
\min_{x_{l}(t), y_{l}(t), \psi_{l}(t)}  \sum_t \Big(\ddot{x}_{l}^2(t)+\ddot{y}_{l}^2(t) + \ddot{\psi}_{l}^2(t) \Big), \label{acc_cost}\\
(x_{l}(t), y_{l}(t), \psi_{l}(t)) \in \mathcal{C}_{b},\label{eq_multiagent}\\
 \dot{x}_{l}^{2}(t) + \dot{y}_{l}^{2}(t)  \leq v^{2}_{max},   
  ~~~
  \ddot{x}_{l}^{2}(t) + \ddot{y}_{l}^{2}(t) \leq a^{2}_{max}, \label{acc_constraint}\\
 \hspace{-0.11cm}
 -\frac{\hspace{-0.07cm}(\hspace{-0.04cm}x_{l}\hspace{-0.04cm}(t)\hspace{-0.07cm}+\hspace{-0.07cm}r_{i}\hspace{-0.04cm}\cos{\psi_{l}\hspace{-0.04cm}(t)}\hspace{-0.06cm}-
 \hspace{-0.06cm}x_{j}\hspace{-0.04cm}(t)\hspace{-0.05cm})^{2}}{a^2}\hspace{-0.11cm}-\hspace{-0.11cm}
 \frac{\hspace{-0.07cm}(\hspace{-0.04cm}y_{l}(t) \hspace{-0.075cm}+\hspace{-0.075cm}r_{i}\hspace{-0.04cm}\sin{\psi_{l}\hspace{-0.04cm}(t)}
 \hspace{-0.075cm}-\hspace{-0.065cm}y_{j}\hspace{-0.04cm}(t)\hspace{-0.06cm})^{2}}{b^2} \nonumber \\
 + 1\leq 0, \forall t, i, j, l\label{coll_multiagent}
 \end{align}
\end{subequations}
\normalsize

\noindent Our cost function (\ref{acc_cost}) minimizes the sum of squared linear and angular accelerations at different time instants. Constraint (\ref{eq_multiagent}) enforces the boundary constraints on the linear and angular positions and their derivatives. \textbf{Note that every problem in the batch has the same boundary conditions.} Inequality (\ref{acc_constraint}) represents the bounds on the total velocity and acceleration. Inequality (\ref{coll_multiagent}) is the standard collision avoidance model wherein a rectangular footprint is approximated through a combination of overlapping circles  (see Fig.\ref{teaser}(a)) \cite{aks_icra20}, \cite{alonso_mora_ad}. The $r_i$'s locates the center of the circle in the robot's local frame. The obstacle's locations are given by $x_j(t), y_j(t)$ and they are assumed to be axis-alligned ellipses with dimensions $(a, b)$.

\noindent \textbf{Existing Works:} Suppose $r_i=0$, i.e., we approximate the footprint of the robot through a single circle. In that case, the optimization has a special form that can be solved efficiently through a technique called convex-concave procedure (CCP) \cite{ccp_boyd}. In comparison, the multi-circle approximation is devoid of such structures due to the presence of orientation variables in (\ref{coll_multiagent}). As a result, one needs to adopt techniques like sequential quadratic programming (SQP) \cite{alonso_mora_ad}, \cite{multi_circle_ad} accessed via frameworks like ACADO \cite{acado}. In the context of reactive one-step planning, works like \cite{aude_billard_convex}, \cite{eorca}  have shown good results for navigating rectangular robots in dynamic environments. However, this setting is slightly orthogonal to our multi-step trajectory optimization set-up that performs planning over long horizons (around 30s).

\noindent \textbf{Connections to Sampling Based Trajectory Optimization:} Local optimization techniques like CCP, SQP rely heavily on the quality of initialization. One way to mitigate this is to use derivative-free sampling-based optimal control \cite{stein_variational_oc}. CEM \cite{cem_1}, \cite{cem_2} is one such method and has recently been combined with local minimizers like gradient-descent \cite{cem_gd}. Authors in \cite{stein_variational_oc} showed that their niche trajectory sampling coupled with gradient-descent could lead to solutions in different homotopies. Our algorithm leads to a similar result but with a very different process. We sample trajectories from the Gaussian distribution proposed in \cite{stomp} but only once and then initialize a powerful multi-convex local minimizer from those samples. 

\noindent \textbf{Note:} Unlike \cite{teb_planner}, \cite{gpmp_homotopy}, works like \cite{stein_variational_oc}, \cite{cem_gd} and ours do not explicitly search for solutions in different homotopy but rely on stochastic sampling to achieve that.

\noindent \textbf{Challenges in Batch Optimization:} Recently, there has been a strong interest in solving several instances of convex quadratic programming (QP) in parallel \cite{opt_net}, \cite{cvxpy_layers}. The core innovation in \cite{opt_net} lies in rewriting the underlying matrix-algebra such that matrices that do not change with the batch index can be isolated, and their factorization can be pre-stored. Our work achieves a similar structure but for a much more challenging constrained non-convex problem. \textbf{Note:} Gradient-descent can be trivially batched but it performed poorly on (\ref{acc_cost})-(\ref{coll_multiagent}) in our experiments.

\section{Main Results}

\subsection{Main Idea}
\noindent Consider the following $l$ equality-constrained QP problems with a special structure that only the vector $\overline{\textbf{q}}_l$ varies across the problem instances.

\vspace{-0.5cm}
\small
\begin{align}
\min_{\boldsymbol{\xi}_{l}} \Big(\frac{1}{2}\boldsymbol{\xi}^{T}_{l}\overline{\textbf{Q}}\boldsymbol{\xi}_{l} + \overline{\textbf{q}}^{T}_{l}\boldsymbol{\xi}_{l}\Big), \qquad
 \text{st: } \overline{\textbf{A}} \boldsymbol{\xi}_{l} = \overline{\textbf{b}} \label{over_1}
\end{align}
\normalsize

The $l^{th}$ optimization problem \eqref{over_1} can be reduced to a set of linear equations as

\vspace{-0.45cm}
\small
\begin{align}
    \begin{bmatrix}
        \overline{\textbf{Q}} & \overline{\textbf{A}}^{T} \\ 
        \overline{\textbf{A}} & \textbf{0}
    \end{bmatrix} \begin{bmatrix}
        \boldsymbol{\xi}_{l}\\ \boldsymbol{\mu}_{l}
    \end{bmatrix} = \begin{bmatrix}
        \overline{\textbf{q}}_{l} \\ \overline{\textbf{b}}
    \end{bmatrix} \label{over_2}
\end{align}
\normalsize

\noindent Where $\boldsymbol{\mu}_l$ is dual optimization variable.  
Since the matrix in \eqref{over_2} is constant for different batch instances, and also all batches are independent of each other, we can compute solution across all the batches in one-shot as follows:

\vspace{-0.35cm}
\small
\begin{align}
\begin{bmatrix} 
\begin{array}{@{}c|c|cc@{}}
\boldsymbol{\xi}_{1} &...& \boldsymbol{\xi}_{l} \\ \boldsymbol{\mu}_{1} &...& \boldsymbol{\mu}_{l}
\end{array}
\end{bmatrix} =
    \overbrace{(\begin{bmatrix}
        \overline{\textbf{Q}} & \overline{\textbf{A}}^{T} \\ 
        \overline{\textbf{A}} & \textbf{0}
    \end{bmatrix}^{-1}) }^{constant}
\begin{bmatrix}
\begin{array}{@{}c|c|cc@{}}
    \overline{\textbf{q}}_{1} & ... & \overline{\textbf{q}}_{l}  \\
    \overline{\textbf{b}} & ... & \overline{\textbf{b}}
    \end{array}\end{bmatrix},
    \label{over_3}
\end{align}
\normalsize
\noindent where, $|$ implies that the columns are stacked horizontally. 

\noindent Our main \textbf{innovation} lies in reformulating the non-convex optimization (\ref{acc_cost})-(\ref{coll_multiagent}) into a form where the most intensive computations have the same structure as (\ref{over_1}) (see (\ref{zeta_1_sepration_3}) and discussions around it).

\subsection{ Reformulating Constraints }

\noindent We extend the polar representation of collision avoidance constraint from \cite {rastgar2020novel, aks_ral21} to a rectangular robot modeled via overlapping circles. Our collision avoidance model has the form $\textbf{f}_c = \textbf{0}$, where
\small
\begin{align}
\forall t,i,j,l  \qquad \textbf{f}_{c}(x_{l}(t),y_{l}(t),\psi_{l}(t)) = \nonumber \\
\hspace{-0.25cm}\left \{ \hspace{-0.23cm}\begin{array}{lcr}
x_{l}(t) \hspace{-0.04cm}+ \hspace{-0.05cm}r_{i} \cos{\psi_{l}}(t)\hspace{-0.07cm}-\hspace{-0.07cm}x_{j}(t) \hspace{-0.07cm}-ad_{ij,l}(t)\cos{\alpha_{ij,l}}(t) \\
y_{l}(t)\hspace{-0.04cm} + \hspace{-0.05cm}r_{i} \sin{\psi_{l}}(t)\hspace{-0.07cm}-\hspace{-0.07cm}y_{j}(t) 
\hspace{-0.05cm}-bd_{ij,l}(t)\sin{\alpha_{ij,l}(t)}
\end{array} \hspace{-0.25cm} \right \}\hspace{-0.10cm},  d_{ij,l}(t)\hspace{-0.10cm} \geq \hspace{-0.07cm}1
\label{collision_avoidance_proposed} 
\end{align}
\normalsize
\noindent where $d_{ij,l}(t)$, $\alpha_{ij,l}(t)$ are the line-of-sight distance and angle between the $i^{th}$ circle of the robot and $j^{th}$ obstacle.
 
Following a similar approach, we reformulate velocity and acceleration bounds in the form
$ \textbf{f}_{v} =\textbf{0} $ and $\textbf{f}_{a} = \textbf{0}$, where

\small
\begin{subequations} 
\begin{align}
 \textbf{f}_{v} =\left \{ \hspace{-0.20cm} \begin{array}{lcr}
 \dot{x}_{l}(t) - d_{v,l}(t)v_{max}\cos{\alpha_{v,l}(t)}  \\
 \dot{y}_{l}(t) - d_{v,l}(t)v_{max}\sin{\alpha_{v,l}(t)} 
\end{array} \hspace{-0.20cm} \right \} 
,d_{v,l}(t) \geq 1, ~  \forall t,l  \label{fv} \\
\textbf{f}_{a} = \left \{ \hspace{-0.20cm}\begin{array}{lcr}
\ddot{x}_{l}(t) - d_{a,l}(t)a_{max}\cos{\alpha_{a,l}(t)} \\
\ddot{y}_{l}(t) - d_{a,l}(t)a_{max}\sin{\alpha_{a,l}(t)}
\end{array} \hspace{-0.20cm} \right \},  d_{a,l}(t) \geq 1, ~\forall t,l \label{fa}
\end{align}
\end{subequations}
 \normalsize

\newtheorem{remark}{Remark}
\begin{remark} \label{convex_sine_cosine}
Our collision avoidance model (\ref{collision_avoidance_proposed}) is convex in the space of $(x_l(t), y_l(t), \cos{\psi}_l(t), \sin{\psi}_l(t))$. 
\end{remark}

\noindent \textbf{Note} that the convexity is with respect to the sine and cosine of $\psi_l(t)$ and not the angle itself.

\subsection{Proposed Reformulated Problem}
\noindent The insights from Remark \ref{convex_sine_cosine} forms the basis of our reformulated trajectory optimization problem. We introduce two slack variables $c_l(t)$, $s_l(t)$ that acts as the copy for $\cos\psi_l(t)$ and $\sin\psi_l(t)$.  Our main trick is to treat $c_l(t)$ and $s_l(t)$ as independent variables and somehow ensure that when the optimization converges, they indeed resemble the cosine and sine of $\psi_l(t)$. To this end, we present the following optimization problem that acts as a substitute for (\ref{acc_cost})-(\ref{coll_multiagent}).

\small
\begin{subequations}
\begin{align}
\min_{\scalebox{0.7}{$\begin{matrix} x_{l}(t), y_{l}(t), \psi_{l}(t), \\
c_{l}(t), s_{l}(t), d_{v,l}(t), d_{a,l}(t), \\d_{ij,l}(t),
\alpha_{v,l}(t),\alpha_{a,l}(t), \alpha_{a_{ij,l}}(t)    \end{matrix}$}}
\sum_t \Big(\ddot{x}^2_{l}(t)+\ddot{y}^2_{l}(t) + \ddot{\psi}^2_{l}(t) \Big)   \label{acc_cost1}\\
(x_{l}(t), y_{l}(t)) \in \mathcal{C}_{b}\label{eq_multiagent1}\\
\textbf{f}_{v}(t) = \textbf{0},  \qquad d_{v,l}(t) \geq 1  \forall t,l \label{cond1}  \\
\textbf{f}_{a}(t) = \textbf{0},  \qquad d_{a,l}(t) \geq 1 \forall t,l \label{fa_fv1} \\
 c_{l}(t) = \cos{\psi_{l}(t)} \label{cos_ref}\\ 
 s_{l}(t) =  \sin{\psi_{l}(t)} \label{sin_ref}\\
\textbf{f}_{c} \hspace{-0.1cm}: \hspace{-0.1cm}\left \{ \begin{array}{lcr}\hspace{-0.2cm}
x_{l}(t) + r_{i} c_{l}(t)-x_{j}(t) -ad_{ij,l}(t)\cos{\alpha_{ij,l}}(t)\\
\hspace{-0.2cm}y_{l}(t) + r_{i} s_{l}(t)-y_{j}(t) 
-bd_{ij,l}(t)\sin{\alpha_{ij,l}(t)} 
\end{array} \right \} \hspace{-0.2cm}\label{coll_con}
 \\ d_{ij,l}(t) \geq 1, \qquad \forall t, i, j,l \label{fc1}
\end{align}
\end{subequations}
\normalsize

\noindent \textbf{Note} how $c_l(t), s_l(t)$ replaces $(\cos\psi_l(t), \sin\psi_l(t))$ in the collision avoidance model. The above optimization is defined in terms of time dependent functions. We convert it into a finite-dimensional form by parametrizing some of the time dependent variables in the following form.

\vspace{-0.2cm}
\small
\begin{equation}
\begin{bmatrix}
x_{l}(t_1)\\
\vdots\\
x_{l}(t_n)
\end{bmatrix} =\textbf{P} \textbf{c}_{x,l}, \begin{bmatrix}
c_{l}(t_1)\\
\vdots\\
c_{l}(t_{n})
\end{bmatrix} = \textbf{P}\textbf{c}_{c,l}, \begin{bmatrix}
s_{l}(t_1)\\
\vdots\\
s_{l}(t_{n})
\end{bmatrix} = \textbf{P}\textbf{c}_{s,l}
\label{param}
\end{equation}
\normalsize
\noindent where, $\textbf{c}_{x,l}, \textbf{c}_{c,l}$ and $\textbf{c}_{s,l}$ are  the  coefficients  of the  polynomials and $\textbf{P}$ is matrix formed with time-dependent basis functions (e.g polynomials). We consider similar expressions for $y_l(t)$ and $\psi_l(t)$ as well in terms of coefficients $\textbf{c}_{y,l}$ and $\textbf{c}_{\psi,l}$, respectively. Note that the derivatives of the variables also are parametrized in a similar way using matrices $\dot{\textbf{P}}, \ddot{\textbf{P}}$.

Using (\ref{param}), we obtain the following finite-dimensional representation of (\ref{acc_cost1})-(\ref{fc1}). 

\vspace{-0.2cm}
\small
\begin{subequations}
\begin{align}
 \min_{\boldsymbol{\xi}_{1,l}, \boldsymbol{\xi}_{2,l}, \boldsymbol{\xi}_{3,l}, \boldsymbol{\xi}_{4,l}} \Big(\frac{1}{2}\boldsymbol{\xi}_{1,l}^T\textbf{Q}\boldsymbol{\xi}_{1,l}  + \frac{1}{2}\boldsymbol{\xi}_{2,l}^{T}\ddot{\textbf{P}}^{T}\ddot{\textbf{P}}\boldsymbol{\xi}_{2,l} \Big)\label{reform_cost},  \\
  \textbf{A}\boldsymbol{\xi}_{1,l} = \textbf{b}, ~~~ \textbf{A}\boldsymbol{\xi}_{2,l} = \textbf{b}_{\psi},   \label{reform_eq}\\
  \textbf{F}\boldsymbol{\xi}_{1,l} = \textbf{g}_{l}(\boldsymbol{\xi}_{2,l},\boldsymbol{\xi}_{3,l}, \boldsymbol{\xi}_{4,l} ) \label{reform_bound},\\
 \boldsymbol{\xi}_{4,l}\geq \textbf{1}. \label{ine_1}
\end{align}
\end{subequations}
\normalsize

\noindent Where, $\boldsymbol{\xi}_{1,l} = (\textbf{c}_{x,l}, \textbf{c}_{c,l}, \textbf{c}_{y,l}, \textbf{c}_{s,l}) $, $\boldsymbol{\xi}_{2,l} = \boldsymbol{c}_{\psi,l} $,
$\boldsymbol{\xi}_{3,l} = (\boldsymbol{\alpha}_{ij,l}, \hspace{0.1cm} \boldsymbol{\alpha}_{v,l}, \hspace{0.1cm} \boldsymbol{\alpha}_{a,l})$, $\boldsymbol{\xi}_{4,l} = (\textbf{d}_{ij,l}, \hspace{0.1cm}\textbf{d}_{v,l},\textbf{d}_{a,l})$ are optimization variables to be obtained. The matrix $\textbf{A}$ are obtained by stacking the first and last rows of $\textbf{P}$ matrix and its derivatives. For details refer to \cite{aks_ral21}. The matrix $\textbf{Q}$ is a block-diagonal stacking of $\ddot{\textbf{P}}^T\ddot{\textbf{P}}$. The matrix $\textbf{F}$ and vector $\textbf{g}_{l}$ are obtained by rewriting constraints  \eqref{cond1} - \eqref{coll_con}  in the following manner.

\small
\begin{align}
\overbrace{ \begin{bmatrix} 
\textbf{A}_{v} & \textbf{0}   \\
\textbf{A}_{a} & \textbf{0}    \\
\textbf{A}_{ob} & \textbf{0} \\
\begin{bmatrix} \textbf{0} & \textbf{P}
\end{bmatrix} & \textbf{0} \\
\textbf{0} &  \textbf{A}_{v}  \\
\textbf{0} &  \textbf{A}_{a} \\
\textbf{0} & \textbf{A}_{ob} \\
\textbf{0} & \begin{bmatrix} \textbf{0} & \textbf{P}
\end{bmatrix}
\end{bmatrix}}^{\textbf{F}}
\begin{bmatrix} \textbf{c}_{x,l} \\ \textbf{c}_{c,l} \\
\textbf{c}_{y,l} \\ \textbf{c}_{s,l}
\end{bmatrix} =
\overbrace{\begin{bmatrix}
\textbf{d}_{v,l}\cos(\boldsymbol{\alpha}_{v,l}) \\
\textbf{d}_{a,l}\cos(\boldsymbol{\alpha}_{a,l}) \\
\textbf{b}_{ob_{1,l}}(\textbf{d}_{ij,l},\boldsymbol{\alpha}_{ij,l}) \\
\cos{\boldsymbol{\psi}_{l}} \\
\textbf{d}_{v,l}\sin(\boldsymbol{\alpha}_{v,l}) \\
\textbf{d}_{a,l}\sin(\boldsymbol{\alpha}_{a,l}) \\
 \textbf{b}_{ob_{2,l}}(\textbf{d}_{ij,l},\boldsymbol{\alpha}_{ij,l}) \\
 \sin{\boldsymbol{\psi}_{l}} 
\end{bmatrix}}^{\textbf{g}_{l}}
\end{align}

\noindent where, 

\vspace{-0.4cm}
\small
\begin{align}
\textbf{A}_{v} = \begin{bmatrix}
\dot{\textbf{P}}  &  \textbf{0}  \\
\end{bmatrix}_{q \times 2n_{v}}  ~
\textbf{A}_{a} = \begin{bmatrix}
\ddot{\textbf{P}}  &  \textbf{0}  \\
\end{bmatrix}_{q \times 2n_{v}}  \nonumber \\
\hspace{-0.15cm}\textbf{b}_{ob_{1,l}} \hspace{-0.1cm}= \hspace{-0.1cm}\begin{bmatrix}
\textbf{x}_{j} \\ \vdots  \\ \textbf{x}_{j} 
\end{bmatrix}_{mq}
\hspace{-0.3cm}+ a \begin{bmatrix}
\textbf{d}_{1j,l} \hspace{0.1 cm}\cos{\boldsymbol{\alpha}_{1j,l}} \\
\vdots\\
\textbf{d}_{mj,l} \hspace{0.1 cm}\cos{\boldsymbol{\alpha}_{mj,l}}
\end{bmatrix}_{mq} \hspace{-0.5cm}, 
\textbf{A}_{ob}\hspace{-0.07cm} = \hspace{-0.07cm}\begin{bmatrix}
\textbf{P}  &  r_{1}\textbf{P}  \\
\vdots  & \vdots  \\
\textbf{P}  & r_{m}\textbf{P}\\
\end{bmatrix}_{mq \times 2n_{v}}\hspace{-1.1cm},
 \nonumber \\
\textbf{b}_{ob_{2,l}} \hspace{-0.1cm}= \hspace{-0.15cm}\begin{bmatrix}
\textbf{y}_{j} \\ \vdots  \\ \textbf{y}_{j} 
\end{bmatrix}_{mq}
\hspace{-0.35 cm}+ b\hspace{-0.1cm} \begin{bmatrix}
\textbf{d}_{1j,l}\sin{\boldsymbol{\alpha}_{1j,l}} \\
\vdots\\
\textbf{d}_{mj,l}\sin{\boldsymbol{\alpha}_{mj,l}}
\end{bmatrix}_{mq}\hspace{-0.5 cm},
\end{align}
\normalsize

\noindent and $\textbf{d}_{v,l}$, $\textbf{d}_{a,l}$, $\textbf{d}_{ij,l}$, $\boldsymbol{\alpha}_{v,l}$, $\boldsymbol{\alpha}_{a,l}$ and $\boldsymbol{\alpha}_{ij,l}$ are constructed by stacking $d_{v,l}(t)$, $d_{a,l}(t)$, $d_{ij,l}(t)$, $\alpha_{v,l}(t)$, $\alpha_{a,l}(t)$ and $\alpha_{ij,l}(t)$ at different time instances.
Similar derivation is used for generating $\textbf{x}_{j}, \textbf{y}_{j}$ and $\boldsymbol{\psi}_{l}$ as well. 

\noindent \textbf{Relaxation:} Till now it is not apparent how our proposed reformulation (\ref{reform_cost})-(\ref{ine_1}) provides any computational gain over (\ref{acc_cost})-(\ref{coll_multiagent}). To make the distinction more explicit, one more layer of simplification is required: we relax the non-convex equality constraints (\ref{reform_bound}) as $l_2$ penalties.

\vspace{-0.35cm}
\small
\begin{align}
 \min_{\boldsymbol{\xi}_{1,l},\boldsymbol{\xi}_{2,l},\boldsymbol{\xi}_{3,l},\boldsymbol{\xi}_{4,l} } 
 \Big(\frac{1}{2}\boldsymbol{\xi}_{1,l}^T\textbf{Q}\boldsymbol{\xi}_{1,l} 
 +\frac{1}{2}\boldsymbol{\xi}_{2,l}^T\ddot{\textbf{P}}^{T}\ddot{\textbf{P}}\boldsymbol{\xi}_{2,l} 
 - \langle\boldsymbol{\lambda}_{l}, \boldsymbol{\xi}_{1,l}\rangle \nonumber \\
  - \langle\boldsymbol{\lambda}_{\psi,l}, \boldsymbol{\xi}_{2,l}\rangle
 +\frac{\rho}{2}\left\Vert \textbf{F} \boldsymbol{\xi}_{1,l}  
 -\textbf{g}_{l}(\boldsymbol{\xi}_{2,l},\boldsymbol{\xi}_{3,l}, \boldsymbol{\xi}_{4,l}) \right \Vert_2^2 \Big) 
 \label{reform_bergman}
 \end{align}
\normalsize

\noindent The vectors ${\lambda}_{l}$ and $\lambda_{\psi,l}$ are known as the Lagrange multipliers and are crucial for ensuring that the $l_2$ penalties of the equality constraints are driven to zero.

\begin{remark}\label{multi_convexity}
For a given $\boldsymbol{\xi}_{2, l}$, and $\boldsymbol{\xi}_{3, l}$ (\ref{reform_bergman}) is convex in the space of $\boldsymbol{\xi}_{1, l}$ and $\boldsymbol{\xi}_{4, l}$. 
\end{remark}

\begin{remark}\label{convex_surrogate}
For a given $\boldsymbol{\xi}_{1, l}$, $\boldsymbol{\xi}_{4, l}$  (\ref{reform_bergman}) is non-convex in the space of $\boldsymbol{\xi}_{2, l}$ but it can be replaced with a convex surrogate from \cite{aks_icra20}. 
\end{remark}

\begin{remark}\label{closed_form_sol}
For a given $\boldsymbol{\xi}_{1, l}$, $\boldsymbol{\xi}_{4, l}$ and $\boldsymbol{\xi}_{2, l}$  (\ref{reform_bergman}) is solvable in closed form for $\boldsymbol{\xi}_{3, l}$ 
\end{remark}

\noindent Remark \ref{multi_convexity}-(\ref{convex_surrogate}) summarizes the multi-convex structure mentioned in the previous sections. 

\subsection{Solution By Alternating Minimization}
\noindent We minimize (\ref{reform_bergman}) subject to (\ref{reform_eq}) through an Alternating Minimization (AM) \cite{jain_mlbook}, wherein at each iteration, only one among $(\boldsymbol{\xi}_{1, l}, \boldsymbol{\xi}_{2, l}, \boldsymbol{\xi}_{3, l}, \boldsymbol{\xi}_{4, l})$ is minimized while the rest are held fixed at values obtained at previous iteration or the previous step of the current iteration. This is summarized in steps 

\begin{enumerate}
\item Initializing ${^{k}}\boldsymbol{\xi}_{2,l},{^{k}}\boldsymbol{\xi}_{3,l}$ and ${^{k}}\boldsymbol{\xi}_{4,l}$ values.
\item Computing ${^{k+1}}\boldsymbol{\xi}_{1,l}$: 
Using AM, \eqref{reform_bergman} with respect to $\boldsymbol{\xi}_{1,l}$ can be rewritten as the following manner

\vspace{-0.4cm}
\small
\begin{align}
{^{k+1}}\boldsymbol{\xi}_{1,l} = \min_{\boldsymbol{\xi}_{1,l}}\hspace{-0.07cm}\Big(\frac{1}{2}\boldsymbol{\xi}_{1,l}^T\textbf{Q}\boldsymbol{\xi}_{1,l} 
 - \langle {^{k}}\boldsymbol{\lambda}_{l}, \boldsymbol{\xi}_{1,l}\rangle \nonumber \\
 +\frac{\rho}{2}\left\Vert \textbf{F}\boldsymbol{\xi}_{1,l}  
 -\textbf{g}_{l}({^{k}}\boldsymbol{\xi}_{2,l},{^{k}}\boldsymbol{\xi}_{3,l}, {^{k}}\boldsymbol{\xi}_{4,l}) \right \Vert_2^2 
  \Big )
  \label{zeta_1_sepration_1}, \textbf{A}\boldsymbol{\xi}_1 = \textbf{b}
 \end{align}
\normalsize

\item Computing ${^{k+1}}\boldsymbol{\xi}_{2,l}$:
Considering updated ${^{k+1}}\boldsymbol{\xi}_{1,l}$, we aim to solve \eqref{reform_bergman} with respect to $\boldsymbol{\xi}_{2,l}$. Thus, we have

\vspace{-0.4cm}
\small
\begin{align}
{^{k+1}}{\boldsymbol{\xi}_{2,l}} = \min_{\boldsymbol{\xi}_{2,l}}\Big( \frac{1}{2}
\boldsymbol{\xi}_{2,l}^T\ddot{\textbf{P}}^{T}\ddot{\textbf{P}}\boldsymbol{\xi}_{2,l}
 - \langle^{k}\boldsymbol{\lambda}_{\psi,l},
\boldsymbol{\xi}_{2,l}\rangle 
 \nonumber \\
 + \frac{\rho}{2}\left\Vert \textbf{F}^{k+1}\boldsymbol{\xi}_{1,l} - 
 \textbf{g}_{l}(\boldsymbol{\xi}_{2,l})\right \Vert_2^2
 \Big), \textbf{A}\boldsymbol{\xi}_{2, l} = \textbf{b}_{\psi} \label{compute_psi_tilta}
\end{align}
\normalsize
\item Computing ${^{k+1}}\boldsymbol{\xi}_{3,l}$:
The optimization problem with respect to $\boldsymbol{\xi}_{3,l}$ is rephrased as:

\small
\begin{align}
\hspace{-0.55cm}{^{k+1}}\boldsymbol{\xi}_{3,l} \hspace{-0.1cm} = \hspace{-0.05cm}\min_{\boldsymbol{\xi}_{3,l}}\hspace{-0.05cm}\Big(\hspace{-0.07cm} \frac{\rho}{2}\left\Vert \textbf{F}^{k+1} \boldsymbol{\xi}_{1,l}  
 -\textbf{g}_{l}(^{k+1}\boldsymbol{\xi}_{2,l}, \boldsymbol{\xi}_{3,l},^{k} \boldsymbol{\xi}_{4,l}) \right \Vert_2^2 \Big) 
 \label{compute_ze_3}
\end{align}
\normalsize
\item Computing ${^{k+1}}\boldsymbol{\xi}_{4,l}$:
In this step, the optimization problem \eqref{reform_bergman} is written with respect to ${^{k+1}}\boldsymbol{\xi}_{4,l}$ as: 

\vspace{-0.3cm}
\small
\begin{align}
\hspace{-0.65cm}{^{k+1}}\boldsymbol{\xi}_{4,l} \hspace{-0.1cm}= \hspace{-0.05cm}\min_{\boldsymbol{\xi}_{4,l}}\Big(\hspace{-0.07cm} \frac{\rho}{2}\left\Vert \textbf{F}^{k+1} \boldsymbol{\xi}_{1,l}  \hspace{-0.05cm}
 -
 \hspace{-0.05cm}\textbf{g}_{l}(^{k+1}\boldsymbol{\xi}_{2,l}, \hspace{-0.2cm}^{k+1}\boldsymbol{\xi}_{3,l}, \boldsymbol{\xi}_{4,l}) \right \Vert_2^2 \Big)
 \label{compute_ze_4}
\end{align}
\normalsize
\item Updating Lagrange multipliers ${^{k+1}}\boldsymbol{\lambda}_{l}$ and ${^{k+1}}\boldsymbol{\lambda}_{\psi,l}$:
\end{enumerate}

\subsection{Analysis and Batch Solution}

\noindent \textbf{Analysing ${^{k+1}}\boldsymbol{\xi}_{1,l}$:}
Optimization problem \eqref{zeta_1_sepration_1} is an equality constrained QP problem with same structure as \eqref{over_1} where

\vspace{-0.3cm}
\small
\begin{align}
\hspace{-0.2cm}\overline{\textbf{Q}}= \hspace{-0.05cm} \textbf{Q} + \rho \textbf{F}^{T}\textbf{F},~~
\overline{\textbf{q}}_{l} = \hspace{-0.05cm} -{^k}\boldsymbol{\lambda}_{l} 
-( \rho\textbf{F}^{T}\textbf{g}_{l}({^k}\boldsymbol{\xi}_{2,l},{^k}\boldsymbol{\xi}_{3,l},{^k} \boldsymbol{\xi}_{4,l}))^{T}
\label{zeta_1_sepration_3}
 \end{align}
\normalsize
Thus solution of \eqref{zeta_1_sepration_1} over all the batches can be computed in one-shot using (\ref{over_3}).

\noindent \textbf{Analysing ${^{k+1}}\boldsymbol{\xi}_{2,l}$:}
By using ${^k}\boldsymbol{\xi}_{1, l}$ obtained in previous step, the optimization problem \eqref{compute_psi_tilta} gets simplified as
\small
\begin{align}
{^{k+1}}{\boldsymbol{\xi}_{2,l}} = \min_{\boldsymbol{\xi}_{2,l}}\Big( \frac{1}{2}
\boldsymbol{\xi}_{2,l}^T\ddot{\textbf{P}}^{T}\ddot{\textbf{P}}\boldsymbol{\xi}_{2,l}
 - \langle ^{k}\boldsymbol{\lambda}_{\psi,l},
\boldsymbol{\xi}_{2,l}\rangle 
 \nonumber \\
 + \frac{\rho}{2}\left\Vert  \begin{bmatrix}
   {^{k+1}}\boldsymbol{c}_{l} \\ {^{k+1}}\boldsymbol{s}_{l}
 \end{bmatrix} - \begin{bmatrix}
    \cos{\textbf{P}\boldsymbol{\xi}_{2,l}}\\
    \sin{\textbf{P}\boldsymbol{\xi}_{2,l}}
 \end{bmatrix}\right \Vert_2^2
 \Big) \label{compute_psi_tilta1}
\end{align}
\normalsize
\noindent Where, $\boldsymbol{c}_{l}$ and $\boldsymbol{s}_{l}$ are constructed by stacking ${c}_{l}(t)$ and ${s}_{l}(t)$ at different time instances. The second non-convex term of \eqref{compute_psi_tilta1} can be replaced with a convex surrogate \cite{aks_icra20}. 
\small
\begin{align}
{^{k+1}}{\boldsymbol{\xi}_{2,l}} = \min_{\boldsymbol{\xi}_{2,l}}\Big( \frac{1}{2}
\boldsymbol{\xi}_{2,l}^T\ddot{\textbf{P}}^{T}\ddot{\textbf{P}}\boldsymbol{\xi}_{2,l}
 - \langle ^{k}\boldsymbol{\lambda}_{\psi,l},
\boldsymbol{\xi}_{2,l}\rangle 
 \nonumber \\
 + \frac{\rho}{2}
\left\Vert
\arctan2({^{k+1}}\boldsymbol{s}_{l},{^{k+1}}\boldsymbol{c}_{l}) - \textbf{P}\boldsymbol{\xi}_{2,l}
\right \Vert_2^2 \Big),~ \textbf{A}\boldsymbol{\xi}_{2, l} = \textbf{b}_{\psi}
\label{compute_convex_surrogate}
\end{align}
\normalsize 
\noindent Optimization (\ref{compute_convex_surrogate}) is again a convex QP of the form (\ref{over_1}) and thus its batch solution can be computed easily through (\ref{over_3}).

\noindent\textbf{Analysing ${^{k+1}}\boldsymbol{\xi}_{3,l}$:} 
We computed and updated $\boldsymbol{x}_{l}$, $\boldsymbol{\dot{x}}_{l}$,  $\boldsymbol{\ddot{x}}_{l}$,  $\boldsymbol{y}_{l}$, $\boldsymbol{\dot{y}}_{l}$ and $\boldsymbol{\ddot{y}}_{l}$ in previous steps. Now, by fixing these updated variables and ${^{k}}\boldsymbol{\xi}_{4,l}$, we can solve the optimization problem \eqref{compute_ze_3}. Note that ${^{k+1}}\boldsymbol{\xi}_{3,l}$ includes three different optimization variables and since each of these variables are independent, we decompose \eqref{compute_ze_3} into three parallel problems as:  
\small
\begin{subequations}
\begin{align}
\forall i, j, \qquad {^{k+1}}\boldsymbol{\alpha}_{ij,l} =   \min_{\boldsymbol{\alpha}_{ij,l}} 
\hspace{-0.08cm}\frac{\rho}{2} \nonumber \\
\times\left\Vert  \begin{matrix}
 \overbrace{  
   {^{k+1}}\textbf{x}_{l}+ r_{i} \cos{{^{k+1}}\boldsymbol{\psi}_{l}} - \textbf{x}_{j}
}^{{^{k+1}}\boldsymbol{\Tilde{x}}_{l}}
 -a{^{k}} \textbf{d}_{ij,l} \cos{\boldsymbol{\alpha}_{ij,l}} 
\\
 \underbrace{
   {^{k+1}}\textbf{y}_{l} + r_{i}\sin{{^{k+1}}\boldsymbol{\psi}_{l}} -  \textbf{y}_{j}
 }_{{^{k+1}}\boldsymbol{\Tilde{y}}_{l}}
 -b{^{k}}\textbf{d}_{ij,l}\sin{\boldsymbol{\alpha}_{ij,l}}
 \end{matrix}
 \right \Vert_2^2 \hspace{-0.2cm}
 \label{compute_zeta_33}  \\
{^{k+1}}\boldsymbol{\alpha}_{v,l} = \min_{\boldsymbol{\alpha}_{v,l}}\frac{\rho}{2} \left\Vert
\begin{matrix}
{^{k+1}}\boldsymbol{\dot{x}}_{l}-{^{k}}\textbf{d}_{v,l}\cos{\boldsymbol{\alpha}_{v,l}}
\\
 {^{k+1}}\boldsymbol{\dot{y}}_{l}-{^{k}}\textbf{d}_{v,l}\sin{\boldsymbol{\alpha}_{v,l}}
 \end{matrix}
\right \Vert_2^2 \label{compute_av1}
\\
{^{k+1}}\boldsymbol{\alpha}_{a,l}  = \min_{\boldsymbol{\alpha}_{v,l}}
\frac{\rho}{2} \left\Vert
\begin{matrix}
{^{k+1}}\boldsymbol{\ddot{x}}_{l}-{^{k}}\textbf{d}_{a,l}\cos{\boldsymbol{\alpha}_{a,l}}
\\
 {^{k+1}}\boldsymbol{\ddot{y}}_{l}-{^{k}}\textbf{d}_{a,l}\sin{\boldsymbol{\alpha}_{a,l}}
 \end{matrix}
\right \Vert_2^2 
\label{compute_acc}
\end{align}
\end{subequations}
\normalsize

\noindent Despite the non-convexity in \eqref{compute_zeta_33}-\eqref{compute_acc}, its solution can be computed easily through some geometric reasoning. We note that each element of $\boldsymbol{\alpha}_{ij}$, $\boldsymbol{\alpha}_{v, l}$, and $\boldsymbol{\alpha}_{a, l}$ are independent of each other for a given position trajectory. Thus,  \eqref{compute_zeta_33} can be seen as a projection of ${{^{k+1}}\boldsymbol{\Tilde{x}}_{l}}$ and ${{^{k+1}}\boldsymbol{\Tilde{y}}_{l}}$ onto an axis-aligned ellipse centered at origin. Similarly, (\ref{compute_av1}) and (\ref{compute_acc})  can be seen as a projection of (${^{k+1}}\boldsymbol{\dot{x}}_{l}$ ,${^{k+1}}\boldsymbol{\dot{y}}_{l}$) and (${^{k+1}}\boldsymbol{\ddot{x}}_{l}$ ,${^{k+1}}\boldsymbol{\ddot{y}}_{l}$) onto a circle centered at origin with radius $\textbf{d}_{v,l}*v_{max}$ and  $\textbf{d}_{a,l}*a_{max}$ respectively. Thus, solutions of  \eqref{compute_zeta_33}-\eqref{compute_acc} can be expressed as 

\vspace{-0.4cm}
\small
\begin{subequations}
\begin{align}
{^{k+1}}\boldsymbol{\alpha}_{ij,l} = \arctan2 ({^{k+1}}\boldsymbol{\Tilde{y}}_{l}, {^{k+1}}\boldsymbol{\Tilde{x}}_{l}).
\label{zeta_2}\\
{^{k+1}}\boldsymbol{\alpha}_{v,l} = \arctan2 ({^{k+1}}\boldsymbol{\dot{y}}_{l}, {^{k+1}}\boldsymbol{\dot{x}}_{l}).
\label{alpha_v}\\
{^{k+1}}\boldsymbol{\alpha}_{a,l} = \arctan2 ({^{k+1}}\boldsymbol{\ddot{y}}_{l}, {^{k+1}}\boldsymbol{\ddot{x}}_{l}).
\label{alpha_a}
 \end{align}
 \end{subequations}
 \normalsize

\noindent \textbf{Analysing ${^{k+1}}\boldsymbol{\xi}_{4,l}$:}
Using updated position and heading trajectory, ${^{k+1}}\boldsymbol{\alpha}_{v,l}$, ${^{k+1}}\boldsymbol{\alpha}_{a,l}$, and ${^{k+1}}\boldsymbol{\alpha}_{ij,l}$ , \eqref{reform_bergman} with respect to $\boldsymbol{\xi}_{4,l}$ reduces to three parallel problems as

\vspace{-0.2cm}
\small
\begin{subequations}
\begin{align}
{^{k+1}}\textbf{d}_{ij,l} = \min_{\textbf{d}_{ij,l} \geq 1 }  \hspace{-0.05cm}\frac{\rho}{2}\left\Vert \begin{matrix}
{^{k+1}}\boldsymbol{\Tilde{x}}_{l}
\hspace{-0.08cm} - a\textbf{d}_{ij,l} \cos{{^{k+1}}\boldsymbol{\alpha}_{ij,l}} \\
{^{k+1}}\boldsymbol{\Tilde{y}}_{l}
 -b\textbf{d}_{ij,l} \sin{{^{k+1}}\boldsymbol{\alpha}_{ij,l}} \end{matrix}  \right \Vert_2^2
 \label{compute_zeta_3} \\
{^{k+1}}\textbf{d}_{v,l} = \min_{\textbf{d}_{v,l}\geq 1}\frac{\rho}{2} \hspace{-0.08cm}\left\Vert \begin{matrix} {^{k+1}}\boldsymbol{\dot{x}}_{l}\hspace{-0.05cm}-\hspace{-0.05cm}\textbf{d}_{v,l}\cos{{^{k+1}}\boldsymbol{\alpha}_{v,l}}\\
 {^{k+1}}\boldsymbol{\dot{y}}-\textbf{d}_{v,l}\sin{{^{k+1}}\boldsymbol{\alpha}_{v,l}}
 \end{matrix}
\right \Vert_2^2  \label{compute_dv}\\
{^{k+1}}\textbf{d}_{a,l} = \hspace{-0.08cm} \min_{\textbf{d}_{a,l}\geq 1}\frac{\rho}{2} \left\Vert \begin{matrix} 
{^{k+1}}\boldsymbol{\ddot{x}}_{l}-\textbf{d}_{a,l}\cos{{^{k+1}}\boldsymbol{\alpha}_{a,l}} \\
  {^{k+1}}\boldsymbol{\ddot{y}}_{l}-\textbf{d}_{a,l}\sin{{^{k+1}}\boldsymbol{\alpha}_{a,l}}
  \end{matrix}
\right \Vert_2^2 \label{compute_da}
\end{align}
\end{subequations}
\normalsize

\noindent QP convex optimization \eqref{compute_zeta_3}-\eqref{compute_da} can be simplified by noting that different time instant elements of $\textbf{d}_{ij,l}$, $\textbf{d}_{v,l}$ and $\textbf{d}_{a,l}$  are independent of each other for a given position and heading trajectory and their derivatives. Thus, these optimization problems reduce to parallel QPs each of which can be solved in symbolic form. Also, each instance in the batch is independent from one another, so all batches can be solved in parallel. As a result,\eqref{compute_zeta_3} optimization problem reduce to $l*m*q$ parallel single-variable QPs. Also, \eqref{compute_dv} -\eqref{compute_da} reduce to $l*q$ parallel QPs. After solving the optimization variables in symbolic form for \eqref{compute_zeta_3}-\eqref{compute_da}, we simply clip them to $[0,1]$ to satisfy the constraints.\\

\noindent \textbf{Lagrange multiplier update:} Lagrange multipliers are updated based on \cite{split_bergman}, \cite{admm_neural} in the following manner.
\small
\begin{subequations}
\begin{align}
{^{k+1}}\boldsymbol{\lambda}_{l} \hspace{-0.1cm}= \hspace{-0.07cm}{^{k}}\boldsymbol{\lambda}_{l} \hspace{-0.08cm}- \rho( \textbf{F}^{k+1} \boldsymbol{\xi}_{1,l}  
 \hspace{-0.07cm}-\hspace{-0.07cm}\textbf{g}_{l}(^{k+1}\boldsymbol{\xi}_{2,l} , ^{k+1}\boldsymbol{\xi}_{3,l},\hspace{-0.04cm}^{k+1}\boldsymbol{\xi}_{4,l}\hspace{-0.03cm} ))\textbf{F}
\label{update_lambda_1} \\
{^{k+1}}\boldsymbol{\lambda}_{\psi,l} \hspace{-0.1cm}= \hspace{-0.05cm}{^{k}}\boldsymbol{\lambda}_{\psi,l} \hspace{-0.07cm} 
-\rho_{\psi}(\arctan2({^{k+1}}\boldsymbol{s}_{l},{^{k+1}}\boldsymbol{c}_{l})-{^{k+1}}\boldsymbol{\psi}_{l})
\textbf{P}\label{update_lambda_psi} 
\end{align}
\end{subequations}
\normalsize
\section{Validation and Benchmarking}
\noindent Due to lack of space, we present the qualitative results in the accompanying video.

\noindent \textbf{Implementation Details:} We used Python to prototype our optimizer and CEM using Jax \cite{jax} as our GPU accelerated linear algebra back-end. Our batch optimizer was always initialized with a Gaussian distribution proposed in \cite{stomp} centered around a straight-line trajectory. We also built an MPC on top of our batch optimizer, wherein we warm-started the Lagrange multipliers $\boldsymbol{\lambda}_l, \boldsymbol{\lambda}_{\psi, l}$ with the solution obtained at the previous control loop. We ran the MPC with a time-budget of $0.04s$ which was enough to perform 10 iterations our optimizer with a batch size of 1000. We used the following benchmarks for validation.

\noindent \textbf{Benchmark 1, Static Crowd:} Here we created an environment with static human crowd of size 30.\\ 
\noindent \textbf{Benchmark 2, Same Direction Flow:} In this benchmark, the human crowd is moving with a max velocity of $0.3 m/s$ and the robot needs to overtake the human crowd while tracking a straight line trajectory with velocity of $1.0 m/s$.\\
\noindent \textbf{Benchmark 2, Opposite Direction Flow:} Here the crowd is moving with a max velocity of $1.0 m/s$ and the robot is moving in the opposite direction tracking a straight-line trajectory with $1.0 m/s$ desired speed.


\noindent\textbf{Metrics:} We use the following metrics for comparison with both base-line MPC and that based on CEM
\begin{itemize}
    \item Success-Rate: Total evaluated problems ($20$ ) in benchmark divided by successful runs with no-collisions.
    \item Tracking Error: defined as $(x_l(t)-x_{des}(t))^2+(y_l(t)-y_{des}(t))^2$. The desired trajectory is chosen as a straight line with a constant velocity.
    \item Acceleration value used in navigation.
\end{itemize}

\subsection{Importance of Multi-circle Approximation}
\noindent Table \ref{circle_multi_compare} establishes the importance of tightly approximating the footprint of the robot by multiple circles. We obtain the presented arc-length and acceleration statistics across 20 problem instances with randomly sampled obstacle configuration. Similar to Fig. 1(a), multi-circle approximation allows navigation in tight spaces and results in, on average $36\%$ decrease in the arc-length as compared to conservative circular-disk approximation of the footprint. The worst-case improvement was even higher at $43\%$. The shorter arc-lengths also translated to lower accelerations with multi-circle approximation, yielding a mean improvement $65\%$.

\begin{table}[h]
\centering
\caption{\small{Acceleration and arc-length comparison for circle vs multi-circle approximation of robot footprint}  }
\label{circle_multi_compare}
\scriptsize
\begin{tabular}{|*{7}{c|}}
\hline
Method & \multicolumn{3}{c|}{Acceleration $[m/s^2]$} & \multicolumn{3}{c|}{Arc-length $[m]$} \\
\cline{2-7}
  & Mean & Min & Max & Mean & Min & Max \\
  \hline
  Multi-circle approx. & 0.99 &0.57 & 1.71 & 19.75& 18.05 & 23.22 \\
  \hline
  Single-circle approx. & 1.64 & 1.19 & 2.28 & 26.93 & 22.28 & 33.41 \\
\hline
\end{tabular}
\normalsize
\vspace{-0.5cm}
\end{table}

\subsection{Comparison with Baseline MPC}
\noindent In this section, we analyze how navigation performance changes with respect to batch size. To this end, MPC that uses our optimizer with batch size one will be referred to as the baseline MPC. This is because it resembles how a typical MPC works where only a single locally optimal trajectory is computed at each control loop. In principle, any off-the-shelf MPC can be used here. We tried using ACADO \cite{acado} but we could not achieve reliable and real-time performance with 30 obstacles. 

\small
\begin{table}[h]
\centering
\caption{\small{Performance Metrics with Respect to Batch Size over all Benchmarks (Mean/Max/Min) } }
\label{baseline_comparison}
\scriptsize
\begin{tabular}{|l|l|l|l|l|}
\hline
Batch Size & Success Rate & Tracking Error $[m]$ & Acceleration $[m/s^2]$\\ \hline
1 (Baseline) & 0.016  & 3.19 / 4.78 / 1.69 & 0.097 / 0.29 / 0.0003\\ \hline
200 & 0.8  & 3.15 / 4.89 / 1.51 & 0.15 / 0.37 / 0.0006  \\ \hline
400  & 0.883  & 3.10 / 4.77 / 1.32 & 0.139  / 0.32 / 0.0005 \\ \hline
600 & 0.9  & 3.09 / 4.74 / 1.39 & 0.139 / 0.32 / 0.0005 \\ \hline
800 & 0.95  & 3.07 / 4.75 / 1.45 & 0.136 / 0.31 / 0.004 \\ \hline
\textbf{1000} & \textbf{0.97}  & \textbf{3.06 /4.65 / 1.56} & \textbf{0.166 / 0.31 / 0.03} \\ \hline
\end{tabular}
\normalsize
\vspace{-0.3cm}
\end{table}
\normalsize
The results are summarized in Table \ref{baseline_comparison}. We show the combined statistics across all the benchmarks. As shown, the baseline MPC could only achieve a meager success rate of 0.016. As the batch size increases, we can see a gradual increase in the success rate. This is one of the core-results of our work that shows the importance of batch optimization. We can also observe a small gradual decreasing trend in tracking error and linear acceleration values with batch size.   
\subsection{Comparison with CEM}
\noindent We now present the most important result of our paper. Table \ref{cem_comparison} shows the comparison of our batch optimizer with CEM, which is considered to be the state-of-the-art derivative-free optimizer and is also extensively used in MPC setting \cite{icem}. Importantly, CEM by construction is equipped to search over different local minima. We heavily vectorized the cost evaluation in CEM and ran it with $8k$ samples, which is $8$ times the batch size of our optimizer. The first row of Table \ref{cem_comparison} shows that our optimizer outperforms CEM in success rate. On benchmark 1, the success rate of our optimizer is 1.0, which is $15\%$ higher than that achieved by CEM. The difference between the approaches increases marginally to $26\%$ on benchmark 2. On benchmark 3, our batch optimizer achieves over two times better success rate than CEM.

Our batch optimizer outperforms in tracking error metric as well. On benchmark 1, our average tracking error is $43\%$ lower. Our best-case tracking error is more than one order of magnitude less than CEM on the same benchmark. The worst-case tracking errors for both ours and CEM are comparable. As we show in the accompanying video, our batch optimizer is able to obtain collision-free trajectories that maneuver in between the obstacles. In contrast, CEM, trajectories take a large detour and are unable to catch up to the desired trajectory. The statistics of min tracking error follow from this behavior. The CEM, however, on average, uses less acceleration value than ours.

\small
\begin{table}[h]
\centering
\caption{\small{Comparison with CEM (Mean/Max/Min)}}
\label{cem_comparison}
\scriptsize
\begin{tabular}{|l|l|l|l|l|}
\hline
Method & Success-Rate & Tracking Error $[m]$ & Acceleration $[m/s^2]$\\ \hline
\textbf{Ours Bench.1} & \textbf{1.0}  & \textbf{2.44 / 5.44 / 0.079} & \textbf{0.13 / 0.32 / 0.0}\\ \hline
\textbf{Ours Bench.2} & \textbf{0.95}  & \textbf{3.11 / 5.01 / 0.88} & \textbf{0.20 / 0.49 / 0.0}  \\ \hline
\textbf{Ours Bench.3}  & \textbf{0.95 } & \textbf{2.74 / 4.82 / 0.07} & \textbf{0.17 / 0.45 / 0.0} \\ \hline
CEM Bench.1 & 0.85  & 3.5 / 5.22 / 1.94  & 0.09 / 0.36 / 0.0 \\ \hline
CEM Bench.2 & 0.75  & 3.99 / 6.87 / 1.12 & 0.09 / 0.28 / 0.0 \\ \hline
CEM Bench.3 & 0.40  & 3.25 / 4.05 / 2.56 & 0.13 / 0.46 / 0.0 \\ \hline
\end{tabular}
\normalsize
\vspace{-0.5cm}
\end{table}
\normalsize

\subsection{Computation Time Scaling}
\noindent Table.\ref{time_comparison} validates the (sub)linear scaling of per-iteration computation time with respect to the number of circles used to approximate the rectangular footprint for a given number of obstacles. The scaling with respect to the number of obstacles also shows a similar trend. We can correlate this trend to the matrix algebra of our optimizer. A linear increase in the number of footprint circles or obstacles leads to a similar increase in the number of rows of $\textbf{F}$ and $\textbf{g}_{l}$ while the number of columns remains constant. Thus, the computation cost of obtaining $\textbf{F}^T\textbf{g}_{l}$ in (\ref{zeta_1_sepration_3}) increases linearly. The $\textbf{F}^T\textbf{F}$ needs to be computed only once since they do not change with iteration or batch index. Similar analysis can be drawn for (\ref{compute_psi_tilta}) as well. Solutions of (\ref{compute_ze_3})-(\ref{compute_ze_4}) are available as symbolic formulae whose evaluation has linear complexity with respect to the number of footprint circles and obstacles.
\small
\begin{table}[h]
\centering
\caption{\small{Time per iteration(ms) for batch size 1000. Solutions typically obtained within $5-10$ iterations.} }
\label{time_comparison}
\scriptsize
\begin{tabular}{|*{8}{c|}}
\hline
 & \multicolumn{7}{c|}{Number of obstacles}\\
 \cline{2-8} 
 & 1 & 5 & 10 & 15 & 20 & 25 & 30 \\
 \hline
 1 circle & 0.46 & 0.5 & 0.58 & 0.76 & 0.88 & 1.02 & 1.12 \\
 \hline
 2 circles & 0.8 & 0.94 & 1.18 & 1.48 & 1.8 & 2 & 2.4 \\
 \hline
 4 circles & 0.74 & 1.22 & 1.82 & 2.4 & 2.8 & 3.6 & 4 \\
\hline
\end{tabular}
\normalsize
\vspace{-0.4cm}
\end{table}
\normalsize

\noindent Table \ref{CPU-GPU} compares the per-iteration computation time of our vectorized batch optimizer over GPUs with multi-threaded CPU implementation. The latter's advantage is that any off-the-shelf optimizer can be run in parallel CPU threads without changing the underlying matrix algebra. On the other hand, CPU parallelization gives rise to the case where every instantiation of the problem competes with one other for the computation resources, slowing down the eventual computation time. Our experimentation showed that a C++ version of our optimizer ran seamlessly for a batch size of 5 with a per-iteration computation time of $0.0015s$. This timing is competitive with our GPU implementation. However, any further increase in the batch size led to a substantial slow down. For example, for a batch size of $6$, the per-iteration CPU time jumped to $0.03s$. Thus, it was more reasonable for larger batch sizes to run as sequential instantiations with mini-batches of 5. We note that a more rigorous implementation might improve the performance of multi-threaded CPU-based batch optimization. However, our current bench-marking establishes the computational benefit derived from several layers of reformulation that induced batch equality constrained QP structure in some critical steps of our optimizer and allowed for effortless GPU acceleration. 

\small
\begin{table}[h]
\centering
\caption{\small{Per-iteration Computation-time for GPU (RTX 3080 laptop) Vs Muti-threaded CPU ($i7, 3.6 GHz, 32 GB$) [s] }}
\label{CPU-GPU}
\scriptsize
\begin{tabular}{|*{7}{c|}}
\hline
& \multicolumn{6}{c|}{Batch Size}\\
 \cline{2-7} 
 & 5 & 200 & 400 & 600 & 800 & 1000 \\
 \hline
 \textbf{GPU} & \textbf{0.0016}& \textbf{0.0017} & \textbf{0.0026} & 
 \textbf{0.0033} & \textbf{0.0039} & \textbf{0.0045} \\
 \hline
 CPU & 0.0015 & 0.075 & 0.12 & 0.18 & 0.24 &  0.3   \\
\hline
\end{tabular}
\normalsize
\vspace{-0.4cm}
\end{table}
\normalsize

\section{Conclusions and Future Work}
In this work, we showed for the first time how we could solve several hundred instances of a challenging non-convex trajectory optimization in parallel and in real-time. Our work has the potential of fundamentally changing how trajectory optimization or MPC are viewed and applied in the context of navigation in dynamic and cluttered environments. Instead of the established norm of computing one locally optimal trajectory, our batch optimizer paves the path for trying out several hundred random initializations of the problem in parallel to explore solutions in different homotopies and escape the local minima trap. In particular, the smooth trajectory sampling of \cite{stomp} gelled particularly well with our optimizer. We showed how MPC based on our batch setting substantially improved the success rate and tracking performance in densely cluttered and dynamic environments compared to state-of-the-art CEM.

Our immediate future efforts are focused on obtaining a deeper understanding of how our batch optimizer can be coupled with sampling-based optimal control with potential applications in reinforcement learning with safety guarantees. We also aim to work on batch trajectory optimization for multiple rectangular holonomic robots.


\bibliography{icra_ral_2022_ref}
\footnotesize{
\bibliographystyle{IEEEtran}
}

\end{document}